\documentclass[runningheads]{llncs}

\usepackage{amsmath,amssymb}
\usepackage{color}

\usepackage{standalone}

\usepackage{microtype}
\usepackage{graphicx}
\usepackage{subcaption}
\usepackage{booktabs} 
\usepackage{listings}
\usepackage{adjustbox}
\usepackage{enumitem}
\usepackage{multirow,bigdelim}
\usepackage[capitalize]{cleveref}

\usepackage{tikz}
\usetikzlibrary{arrows.meta}

\begin{document}
\pagestyle{headings}
\mainmatter

\def\ACCV20SubNumber{0422}

\title{Real-Time Segmentation Networks should be Latency Aware}

\titlerunning{Real-Time Segmentation Networks should be Latency Aware}

\authorrunning{E. Courdier et al.}

\author{Evann Courdier\inst{1,2} \orcidID{0000-0003-2135-1910}, Fran\c{c}ois Fleuret\inst{2,3} \orcidID{0000-0001-9457-7393} \thanks{Work done when F. Fleuret was at Idiap Research Institute}}

\institute{Idiap Research Institute, Martigny, Switzerland\\
\email{evann.courdier@idiap.ch}\and
EPFL, Lausanne, Switzerland \and
University of Geneva, Geneva, Switzerland\\
\email{francois.fleuret@unige.ch}}

\maketitle

\begin{abstract}

As scene segmentation systems reach visually accurate results, many recent papers focus on making these network architectures faster, smaller and more efficient. In particular, studies often aim at designing `real-time' systems. Achieving this goal is particularly relevant in the context of real-time video understanding for autonomous vehicles, and robots.

In this paper, we argue that the commonly used performance metric of mean Intersection over Union (mIoU) does not fully capture the information required to estimate the true performance of these networks when they operate in `real-time'.
We propose a change of objective in the segmentation task, and its associated metric that encapsulates this missing information in the following way: We
propose to predict the future output segmentation map that will match the {\it future} input frame at the time when the network finishes the processing.
We introduce the associated latency-aware metric, from which we can determine a ranking.

We perform latency timing experiments of some recent networks on different hardware and assess the performances of these networks on our proposed task. We propose improvements to scene segmentation networks to better perform on our task by using multi-frames input and increasing capacity in the initial convolutional layers.

\end{abstract}

\section{Introduction}

Recent image segmentation networks achieve near-human level performance due to their expressive power, and more focus is on designing architectures that are faster, and can run on smaller hardware with less memory and computing power. In particular, enabling real-time segmentation is critical for applications in robotics, autonomous driving or medical imaging during surgery.

The primary way currently used to assess performance is a task whose objective is the prediction of the input frame's segmentation, which is compared to the input frame's ground-truth segmentation using a given metric (\textit{e.g.} mIoU). In what follows, we will use `accuracy' to refer to such a metric.
For networks aiming at low-latency, researchers also estimate efficiency with the \textit{Frames Per Second} (FPS) metric, or its inverse the Seconds Per Frame metric, also called \textit{latency}.

On real-time segmentation benchmarks, networks are ranked according either to some accuracy metric or latency.
Often, accuracy-latency charts also allows to quickly estimate a new network overall performances.
However, we claim there still is critical information missing to the practitioner: What is the actual accuracy of the system when deployed and used in practice? In other terms, we want to help answer the question of how the system's latency will affect the relevance of its predictions.

We propose an intuitive extension to the usual video segmentation task by introducing a change in the objective. We change the goal from predicting input frame segmentation to predicting future frame segmentation. Going beyond introducing a useful metric, our `latency-aware' task aims at encouraging researchers to focus on a more relevant goal for real-time contexts, i.e. designing \textit{anticipatory} networks.

The change we propose in the objective definition is straightforwardly applicable to a wide range of problem domains ({\it e.g.} object tracking, object detection, object segmentation, pose estimation). In the remainder of this paper, we will focus on the scene semantic segmentation task and perform our experiments on it.

Our contributions are as follows:
\begin{itemize}[itemsep=2pt,topsep=2pt]
\item We propose a simple, and relevant task that aims to assess actual performance of real-time networks,
\item we highlight the associated metric and discuss its benefits,
\item we analyse the relevance of the metric through multiple experiments on different scene segmentation networks,
\item we propose improvements to a fast image-segmentation network for better performance on our task by taking multiple frames as input and increasing the number of channels of early convolutional layers.
\end{itemize}

We will make our code publicly available at the time of the conference.


\section{Related Work}

\subsection{Image Semantic Segmentation}

Most popular approaches for tackling Semantic Segmentation use a variant of powerful deep classification networks that are made fully convolutional, with all final fully connected layers replaced by convolutions. That seminal idea is at the core of the FCN paper \cite{long2015fully}.

The main issue coming with this technique is that it significantly reduces the image resolution in order to retrieve semantic information. Subsequent models for semantic segmentation are built as a ``fully convolutional network'' and attempt to cope with the dimension reductions, while increasing the Receptive Field.

One commonly used techniques is to use a \textit{decoder network} plugged after the FCN to upsample the segmentation map using transposed convolution, as first did \cite{ronneberger2015u} and \cite{badrinarayanan2017segnet} with SegNet and U-Net. This setup allows to merge spatially rich shallow layers into semantically rich deeper layers.

DeepLab v2\cite{chen2017deeplab} later proposed to use \textit{dilated convolutions} \cite{yu2017dilated} to avoid downsampling. This allows to process images with a large field of view without having to reduce them, but it comes with a larger computational complexity.

Finally, \cite{zhao_pyramid_2016} proposed to use a ``\textit{Spatial Pyramidal Pooling}'' (SPP) module \cite{he2015spatial} for segmentation. SPP pools the image simultaneously at different resolutions over a grid, therefore enlarging the Receptive Field. This allows to incorporate a larger context, and take into account higher-level semantic.

Many works followed with techniques to produce high-quality segmentation \cite{tang2018regularized,takikawa2019gated,zhu2019improving,valada2019self}, including better ways to extract features \cite{peng2017large,zhang2018exfuse,he2019dynamic,wu2019fastfcn} and to take into account context \cite{ding2018context,yuan2019object}. Some recent works also proposed attention-based methods \cite{li2019attention,fu2019dual,huang2019ccnet,tao2020hierarchical,zhang2020resnest,choi2020cars} and neural architecture search for image segmentation \cite{liu2019auto,zhang2020dcnas,nekrasov2020architecture}.

On a different application domain, similarly to the change we propose, backbones with enlarged front-end have been used with success for object detection \cite{zhu2019scratchdet} were the authors are training their network without ImageNet pretraining.

\subsection{Real-time Semantic Segmentation}

Reducing the computational cost and the memory cost of deep segmentation systems is critical for many applications that need to run real-time on slow hardware. 
A precursor in fast segmentation is ICNet \cite{zhao_icnet_2017}, which is a fast network that uses multi-scale processing with a special fuse block to merge multi-scale information.

One way of optimising neural network architecture for speed is by using factorised convolutional blocks, \textit{e.g.} factorizing kernels $k \times k$ into $1 \times k$ and $k \times 1$ kernels as does ERFNet \cite{romera2017erfnet}. It can also be achieved using group convolutions, and methods such as ShuffleNet \cite{zhang2018shufflenet} propose different ways to create connections between groups.

One can also use depthwise separable convolution (DSC), which are the combination of depthwise and pointwise convolutions. These DSC are used to lower the number of parameters and makes the inference faster, at the cost of accuracy. They are used broadly in MobileNets \cite{howard_mobilenets:_2017,sandler2018mobilenetv2}.

Another important idea of these networks is to quickly downsample images in order to perform most of the processing at a smaller resolution and avoid full resolution processing. This idea is key to the design of ENet \cite{paszke2016enet}.

BiSeNet and BisenetV2 \cite{yu_bisenet:_2018,yu2020bisenet} proposed a way to separate the localization problem from the semantic extraction problem, and then to merge the two information appropriately.

Recent work such as FasterSeg network \cite{chen2019fasterseg} also use Neural Architecture Search to succesfully discover fast neural architectures for semantic segmentation.

Among fast segmentation networks, Swiftnet \cite{orsic2019defense} is another recent work that proposes an architecture with a light-weight ImageNet-pretrained Resnet followed by a simple decoder using lateral connections similarly to U-Net.
For our work, we choose SwiftNet as one of our base networks for its simplicity and its speed.

\subsection{Video Scene Segmentation Networks}

Another part of the literature focuses on designing \textit{video} segmentation systems. More specifically, these works try to leverage the temporal correlation of consecutive frames in a video to improve the next-frame prediction and reduce computation and latency. However, most works in this domain are more focused on improving segmentation accuracy than reducing the latency.

The Clockwork net in \cite{shelhamer_clockwork_2016} is a model that leverages temporal correlation by running different parts of the network at each time-step conditionally to how much the video has changed from the previous frame. This technique has the disadvantage of not providing a fixed frame-rate.

Another direction to address the problem is to try propagating previous features to consecutive frames to avoid recomputing very similar features for following frames, as is done in \cite{zhu_deep_2016}, even though their design is not meant for real-time.

The work \cite{li2018low} built on these two previous ideas. Their network decides at each frame whether to propagate previous features or to recompute the entire segmentation map. They improved the clockwork design to reduce the maximum latency but did not reach real-time.

Other works use predictive learning, that is predicting future frames or flow motion using past frames and segmentations to help current segmentation \cite{luc2017predicting,jin2017video,jin2017predicting}.

Video temporal coherence is also used along with representation warping to produce better future segmentation maps. Warping is either applied at the feature level \cite{saric2020warp,gadde_semantic_2017} or directly at the segmentation map level \cite{terwilliger2019recurrent}, and possibly combined with existing features to produce the output.
However, these works are not focused on time efficiency.

A Bayesian approach for multi-modal future prediction of scene segmentation was also proposed in \cite{bhattacharyya2018bayesian} that allows to take into account model and observation uncertainty.

Recently, Temporally Distributed Network \cite{hu2020temporally} was introduced for fast video segmentation. It uses a teacher-student design where fast student networks have to predict - in turns - part of the feature map of the teacher network.

\section{A new task for real-time networks}

\subsection{Motivation for latency awareness}

Real-time network performance is usually assessed through accuracy-latency charts that help in understanding a network's trade-offs.
These charts provide \textit{instant accuracy} of networks, \textit{i.e.} the accuracy between the network's prediction and the input's ground truth.
In practice however, networks may need a few seconds before they make a prediction. During that time, the scene has changed and the network prediction does not match that change. It is then particularly useful to compare the network prediction with this new scene's segmentation.
This important comparison is missing from latency-accuracy charts as they do not provide the actual accuracy (compensated for time-delay) that one will get in practice. More, it does not provide neither a total order relation nor a ranking to compare various real-time networks, as can be seen on benchmark websites such as \textit{paperswithcode.com}\footnote{\url{https://paperswithcode.com/task/video-object-segmentation}}, 
We believe that it is therefore relevant to introduce a new objective for the segmentation task that takes into account network latency. This allows to get a meaningful accuracy information and practical benchmarking of networks.

\subsection{Defining a new objective for real-time networks}

We propose to change the objective of the segmentation task: currently, the objective of the task is to predict the input frame segmentation. Instead, we propose as objective to predict the segmentation of the ``future'' frame \textit{at the time the network finishes its computation}.

Formally, let us consider a video sequence and let $I_t$ and $S_t$ denote respectively the frame at time $t$ and its ground truth segmentation. Let $F$ denote the operation of a network (say semantic segmentation) that takes $l_F$ milliseconds to process the current input $I_t$.
The common objective is to improve the metric:
\begin{equation}\text{acc}\left(F(I_t), S_t\right)\end{equation}
while our task proposes to optimise for:
\begin{equation}\text{acc}\left(F(I_t), S_{t + l_F}\right)\label{eq_metric}\end{equation}

Instead of predicting the segmentation of the input frame, our task expects systems to predict the segmentation of a future frame, thus acknowledging the network latency.

This objective is particularly relevant for real-time applications in which we are usually more interested in what is currently happening than in what was a few seconds before : it is useful compare the information we get at a given time $t$ using a network $\left(F(I_{t-l_F})\right)$ with the information we ideally would like to get at that time $\left(S_{t}\right)$.

As said earlier, this change of objective is applicable to a whole range of different tasks
such as object segmentation, object detection, object tracking, pose detection, etc.
We focus on the scene segmentation task for this work.

\begin{figure*}[ht]
\vskip 0.2in
\begin{center}
\centerline{\includegraphics[width=\textwidth]{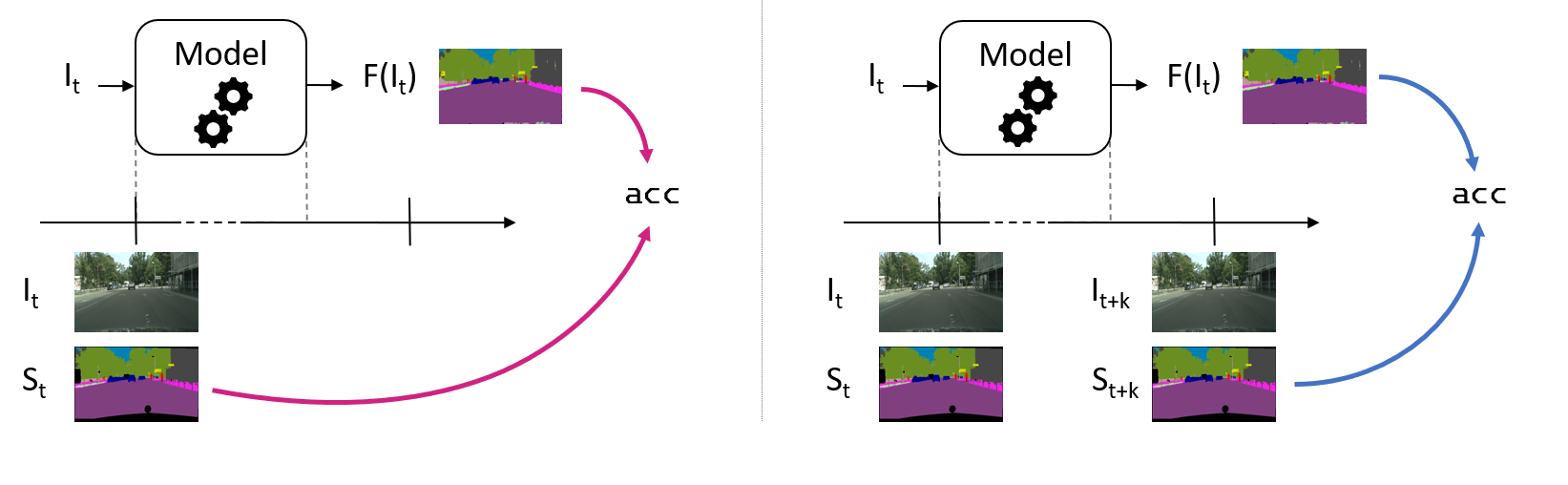}}
\caption{Left: How mIoU is computed now; the output of the network is compared to the ground truth segmentation of the \textit{image in input of the network}. Right: Our proposed way to measure mIoU; the output of the network is compared to the ground truth segmentation of the \textit{future image} when network finishes processing.}
\label{schema}
\end{center}
\vskip -0.2in
\end{figure*}

\subsection{Corresponding Latency-Aware Metric}

For scene semantic segmentation, the metric commonly used is the mean Intersection over Union (mIoU).
Our new task naturally defines a metric that depends on the latency of the network. We term it ``Latency-Aware mIoU" (LAmIoU), which is defined as per \cref{eq_metric}.\\
Considering this metric is interesting as it carries an additional practical meaning compared to the classical instant mIoU: the accuracy (LAmIoU) that this metric outputs is the accuracy that one will see in practice when running this network in a real-time setting on the given hardware device.

\subsection{Use with video datasets}

In practice, a video sequence is collected with a specific sampling frequency (there is some time delay $d$ between two sampled frames), and thus the dataset does not have a frame for every time $t$. For our task we therefore use the frame sampled just \textit{after} the model has output a prediction as shown in \cref{frame_used}.

More precisely, let's assume that $t = 0$ when frame of index $0$ enters the network and consider a video sequence with a delay $d$ between two frames (fps $ = 1/d$). Then, the index of the segmentation map that the metric would use as ground truth is: 
\begin{equation}\label{eq:1}
k_F = \lceil l_F / d \rceil
\end{equation}

In what follows, when we refer to $t + k_F \times d$, we will simplify notation and write $t+k_F$.

\begin{figure}[h]
\centering
\begin{tikzpicture}[font=\scriptsize]

  \draw[-{Triangle[scale=1.5]}] (-0.5, 0) -- (4.5, 0);

  \draw[shorten >=2pt,shorten <=2pt,{Triangle[scale=1.5]}-{Triangle[scale=1.5]}] (0, -1) -- (2.3, -1) node[midway,below] {$l_f$};

  \foreach \x in { 1, 2, 3, 4 }
    \draw (\x, -0.1) -- ++(0, 0.2);

  \draw[dashed] (0, 0.5) -- ++(0, -2.5) node[below,align=center] {Start \\ processing};
  \draw[dashed] (2.3, 0.5) -- ++(0, -2.5) node[below,align=center] {End \\ processing};

  \node[fill=white,inner sep=3pt] at (0, -0.35) {$t$};
  \node[fill=white,inner sep=3pt] at (3, -0.35) {$t+k_f$};

  \draw[{Triangle[scale=1.5]}-] (3, -0.6) |- ++(0, -1) -- ++(1, 0) node[right,align=center] {Target frame \\ used for the \\ task};

  \draw[rounded corners=3pt] (0., 0.4) rectangle (2.3, 1.2) node[pos=0.5] {Model};

\end{tikzpicture}
\caption{Target frame used for our proposed objective}
\label{frame_used}
\end{figure}

Note that the performance on this task is hardware-dependent.
Indeed, as network latency depends on the device, the frame used for metric computation depends on it as well.
Knowing the value of this metric for a network on multiple hardware allows one to pick the right network and hardware depending on precision needs.

\section{Dataset, Models and Experimental setup}

\subsection{Dataset}
\label{dataset}

Dense pixel-level manual annotation of videos for scene segmentation is not feasible due to the time and economic costs involved. Cityscapes dataset \cite{cordts2016cityscapes} was estimated to take about 90 minutes per-frame for annotation and verification, and thus only provides sparse annotations of one frame per video sequence. 
We choose this dataset to conduct our experiments as it contains video sequences and its use is widespread as a segmentation benchmark.

This dataset contains 2,975 training, 500 validation and 1,525 testing video sequences.
Each sequence contains 30 frames, and the $20^{th}$ frame is annotated with fine pixel-level class labels for 19 object categories.
A sequence is 1.8s long, therefore the frame rate is approximately 16.6 fps and there is about 60ms between each frame.
As Cityscapes contains only the ground truth segmentation for one image per sequence, we process as follows:

\begin{enumerate}
\item We time the latency $l_F$ of the network
\item We determine how many frames of offset $k_F$ this time corresponds to: $k_F = \lceil l_F / 0.06 \rceil$ (0.06 = 60ms)
\item We use as input of the network the frame of index $20 - k_F$, as we only have the $20^{th}$ frame's ground truth segmentation
\end{enumerate}

Note that the Cityscapes framerate is about half of the one usually encountered in videos. This may be slightly detrimental as the higher the framerate is, the more precise the latency-aware metric will be.

\subsection{Networks}
\label{net_desc}

For our experiments, we choose 
two image segmentation networks: SwiftNet \cite{orsic2019defense} and DeepLab-V3+ \cite{chen_encoder-decoder_2018} with 2 different encoders : ResNet-101 and MobileNet v2 \cite{sandler2018mobilenetv2}.

\subsubsection{SwiftNet}

SwiftNet is a state-of-the-art network in real-time segmentation.
For our experiments, we build this network as detailed in the original paper and describe it below.
It is a network with an encoder-decoder structure:

\begin{itemize}
\item The encoder backbone is a classical ResNet-18 whose fully connected layers have been removed to make it fully convolutional.
\item A Spatial Pyramidal Pooling Module with 4 different pooling layers of grid size (1,2,4,8) is plugged in output of the encoder to increase its receptive field.
\item Finally, a decoder with 3 upsampling modules recovers original image resolution. An upsampling module upsamples the previous layer's output and then merges it with a skip connection coming from the encoder.
\end{itemize}

We will refer to it as {\bf SwiftNet-R18}. It has approximately 12M parameters. On Cityscapes validation set, it reaches 75\% mIoU and runs at about 40 fps on a GTX 1080 Ti.
On this hardware, SwiftNet has a latency of 26 ms. This means we have to use $k_F = \lceil 26 / 60 \rceil = 1$ frame offset to compute the latency-aware mIoU.

\subsubsection{DeepLab v3+}

DeepLab v3+ is a state of the art network for image segmentation.
It has an encoder-decoder architecture very similar to that of SwiftNet:
\begin{itemize}
    \item We use two different encoder backbones :
    \begin{itemize}
        \item A dilated ResNet-101 network stripped of its fully connected layers. We use an output stride of 16, so the last two residual blocks make use of dilated convolutions to enlarge the receptive field.
        \item A MobileNet-V2 network as described in \cite{sandler2018mobilenetv2}. It uses depthwise separable convolutions within ``inverted'' residuals blocks separated by linear bottlenecks.
    \end{itemize}
    \item An Atrous Spatial Pyramidal Pooling module is plugged after this encoder. It convolves the encoder output with 4 atrous convolutions using different dilation rates: (1, 6, 12, 18).
    \item Finally, a small decoder upsamples the ASPP output and concatenates it with low-level features from the encoder. The decoder blends them with a convolution and upsamples the output to the original input image size.
\end{itemize}

When using a ResNet-101 as backbone, the whole network has approximately 60M parameters and reaches 77\% mIoU on Cityscapes validation set and runs at 5 fps. We will refer to it as {\bf DeepLab-R101}. On our hardware, it has a latency of 195 ms, which means we have to use $k_F = \lceil 195 / 60 \rceil = 4$ frame offsets to compute the latency-aware mIoU.

When using a MobileNet backbone, the model has about 5.5M parameters. It reaches 72\% mIoU on Cityscapes and runs at 13 fps. We will refer to it as {\bf DeepLab-MN.}
It has a latency of 76 ms, so we have to use $k_F = \lceil 76 / 60 \rceil = 2$ frame offsets.

\subsection{Adapting SwiftNet for our task}
\label{sw_pres}

As we will discuss further, it is useful to input previous frames along with the current frame when training to predict a future segmentation map (that corresponds to the latency-aware objective). When we added more input channels to the first layer of the network, we noticed it is beneficial to increase the initial layers capacity by enlarging the number of channels.
We construct a variant of SwiftNet that takes multiple frames in input, and where we expand the number of kernels in the initial convolution layers to deal with the increased input size.
Specifically, in the case of two input frames, we replace the first layer: 
\[
\text{conv}(3, 64, 7 \times 7, s=2)
\]
with the following block of four layers:
\[
\begin{array}{l}
  \text{conv}(6, 130, 7\times7, s=2)\\
  \text{BN}(130)\\
  \text{ReLU}\\
  \text{conv}(130, 64, 3 \times 3, s=1)
\end{array}
\]

Note that we cannot introduce changes affecting multiple encoder layers as this would prevent us from reusing pretrained weights for the ResNet encoder, which represents SwiftNet's main strength. We have experimented with non-pretrained ResNet and observed a 6\% mIoU drop on average.

The newly created convolutions were initialized using He's initialization \cite{he2015delving} rule.
We will refer to our extended SwiftNet version as {\bf SwiftNet-R18-X}.

\subsection{Training}
All experiments are performed with the PyTorch framework.
We use ImageNet-1k pretrained weights for all encoders in our networks.

\subsubsection{Data augmentation}
We use image crops of $768 \times 768$. We do standard image augmentation with random horizontal flip, random scaling from 0.75 to 1.5 and random Gaussian blur.

\subsubsection{SwiftNet}
For SwiftNet, we use a batch size of $12$ and train using the Adam optimiser with default parameters. We use a learning rate of $5 e{-4}$ and a weight decay of $1 e{-4}$. We also set a smaller learning rate of $1 e{-4}$ for the part that is ImageNet-pretrained. We train the network for $200$ epochs and use a cosine annealing schedule with $\eta_{min}=1e{-6}$.

\subsubsection{DeepLab v3+}
For DeepLab, we use a batch size of $10$ and train using the SGD optimiser with momentum of $0.9$. We use a learning rate of $5 e{-2}$ and a weight decay of $5 e{-4}$. We similarly set a smaller learning rate of $5 e{-3}$ for the part that is ImageNet-pretrained. We train the network for $200$ epochs and use a poly schedule with a power of 3.

\section{Experiments and Results}

We perform experiments to evaluate the effect of network latency on the LAmIoU and to understand the changes in the training to suit the proposed objective.

\subsection{Effect of latency on the LAmIoU}

The experiments described in this subsection are performed with \textit{DeepLab-R101} and \textit{SwiftNet-R18}. These two networks are modified to accept in input 2 frames $X_{t-1}, X_t$, as will be detailed later.

\subsubsection{Decay of the LAmIoU with the frame offset}

\begin{figure}[ht]
\centering
\includegraphics[width=0.95\linewidth]{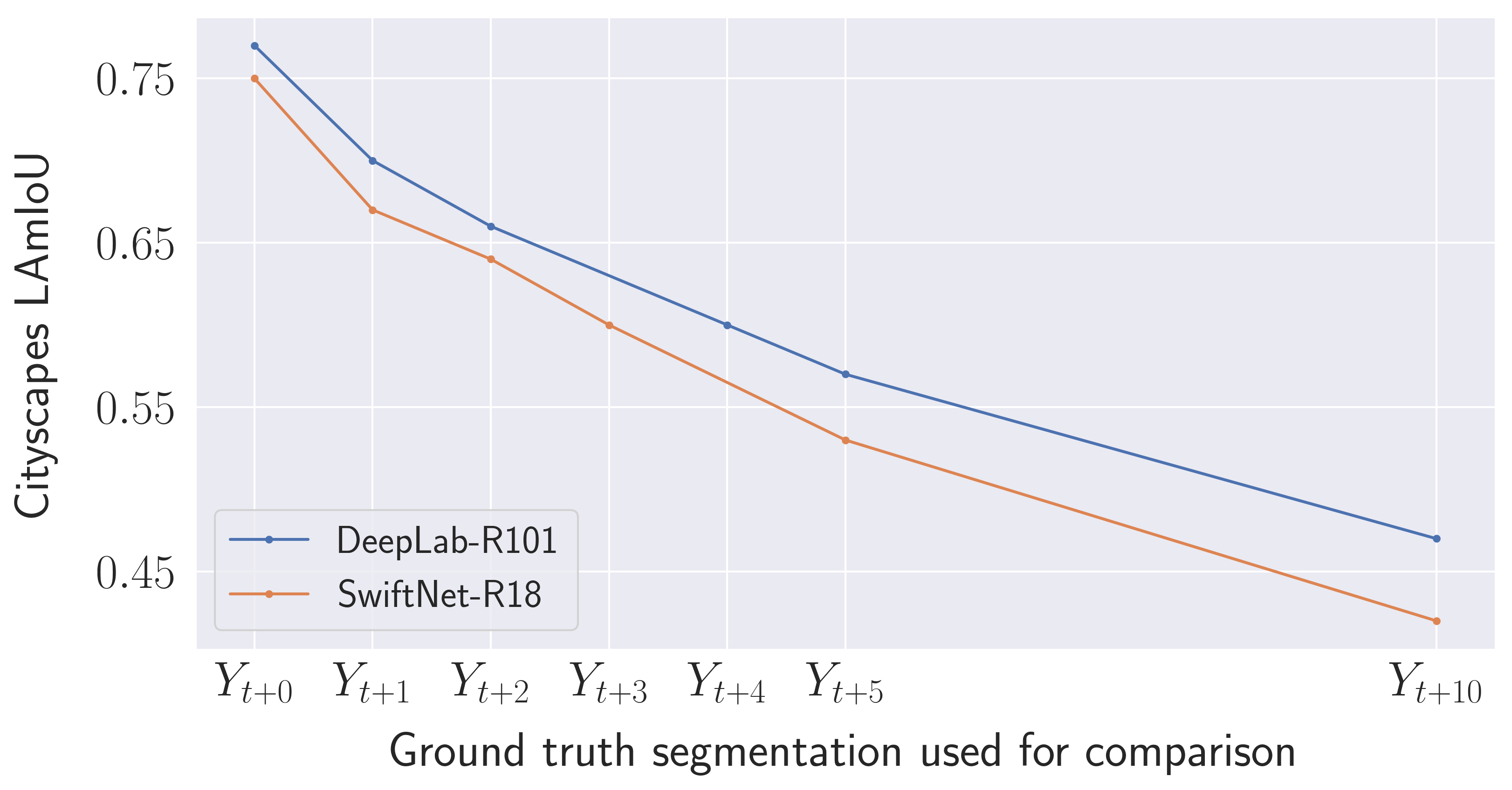}
\caption{LAmIoU decay with different offsets in the objective on Cityscapes validation set}
\label{decay_target}
\end{figure}

In \cref{decay_target}, we plot the LAmIoU vs frame offsets. The two networks considered here are both trained and evaluated on the future segmentation map (with offset).

We notice that the LAmIoU drops quickly and the decrease is consistent between networks: an offset of 2 frames is enough to lose 10\% mIoU for both networks on Cityscapes. In practice, we can expect this order of magnitude of mIoU drop, depending on the hardware.

\subsubsection{Decay of the LAmIoU with the hardware}

Each network has a different latency per hardware. Therefore, the frame offset used for training and computing the metric is also different per hardware.

\begin{figure}[ht]
\centering
\includegraphics[width=.95\linewidth]{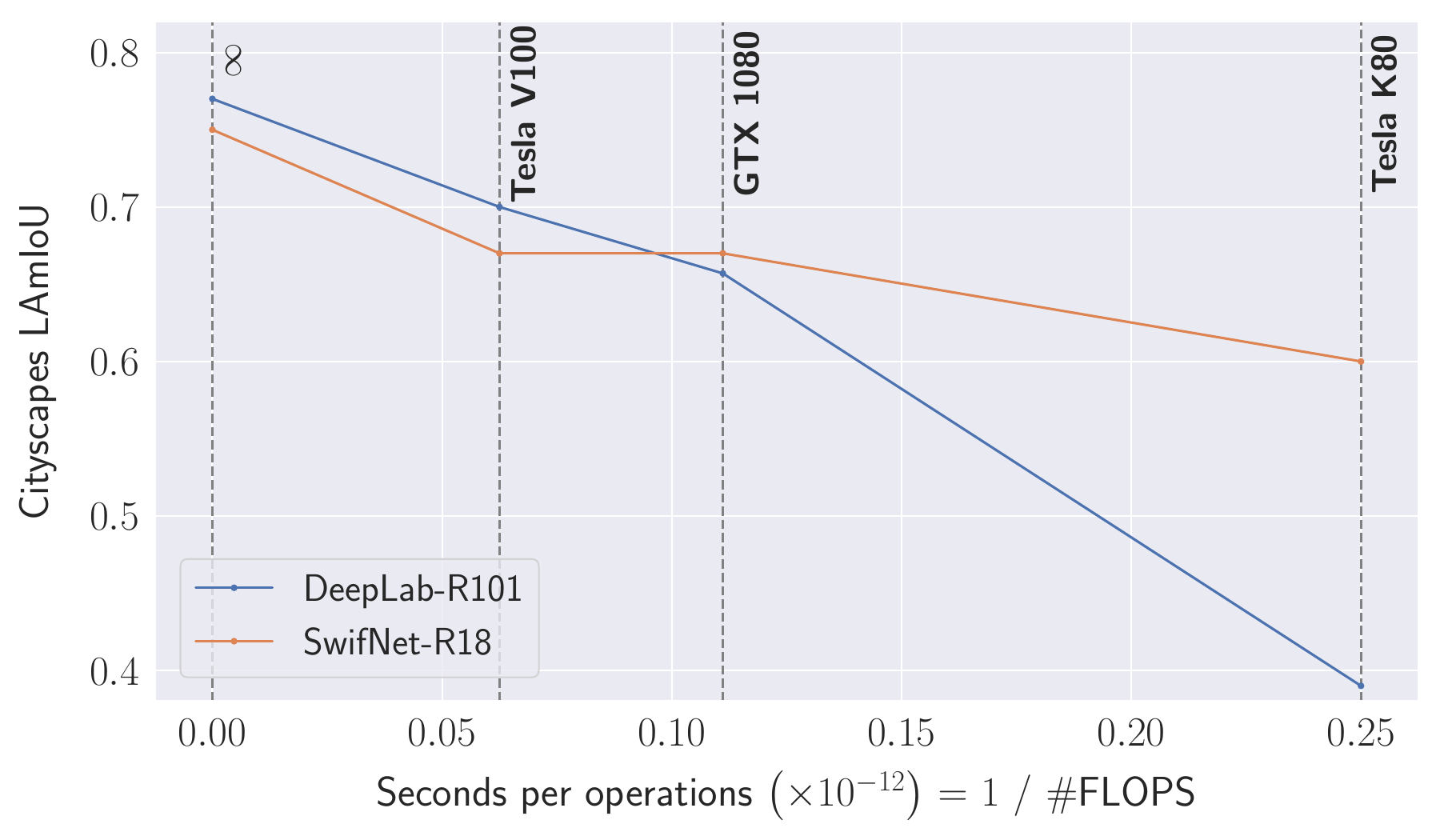}
\caption{LAmIoU decay with hardware speed on Cityscapes validation set}
\label{decay_hardware}
\end{figure}

We perform timing experiments on \textit{Tesla V100}, \textit{GTX 1080 Ti} and \textit{Tesla K80} GPUs for our two networks. We estimate latencies on these hardware and then train the networks with the corresponding frame offsets. In \cref{decay_hardware}, we report the LAmIoU with respect to the inverse hardware speed (inverse of flops).

This plot exhibits an interesting and foreseeable fact: the slower deeplab network, whose ``instant'' mIoU is higher, performs worse than the faster SwiftNet network on slow hardware when measuring the LAmIoU.
This graph illustrates the need for a latency aware metric in real-time contexts, which allows for a simple and fairer comparison of networks.

\subsection{Optimising network training for our latency-aware objective}

In this subsection, we investigate how changing the inputs and training target affects the LAmIoU.
Precisely, we train using three different configurations and comment on the differences in the networks output.
These experiments are performed on a \textit{GTX 1080 Ti} GPU for each of the four networks described in \cref{net_desc}.

\subsubsection{First configuration}

First, we evaluate the four networks when trained with the usual objective (input $I_t$ and target $S_t$) but evaluated with LAmIoU.
The results are reported in second line of \cref{bigtable}.
Compared to their instant mIoU, we notice a significant drop between 10\% and 30 \%.

\subsubsection{Second configuration}

We train the networks for the proposed objective of predicting the future segmentation ground-truth used by the LAmIoU (input $I_t$ and target $S_{t+k}$). The value of $k$ is different for each network since each has a different latency. Therefore, they are trained and evaluated with a differently offset ground-truth.

\begin{figure*}[ht]
\centering
    \begin{subfigure}{0.33\textwidth}
        \centering
        \includegraphics[width=0.85\linewidth]{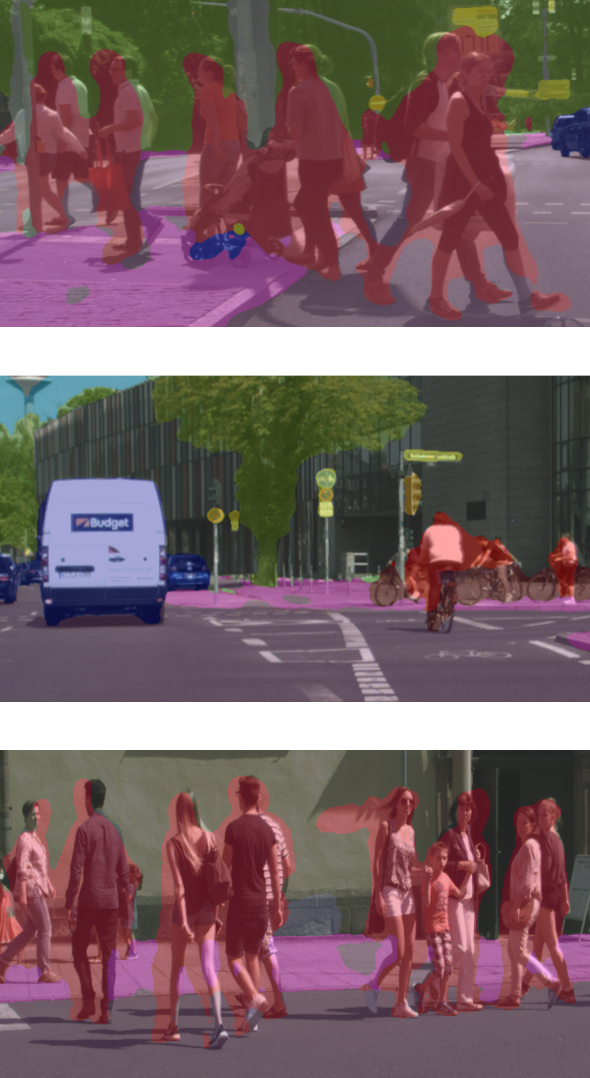}
        \caption{$I_t$ and $F(I_t)$}
    \end{subfigure}%
    \begin{subfigure}{0.33\textwidth}
        \centering
        \includegraphics[width=0.85\linewidth]{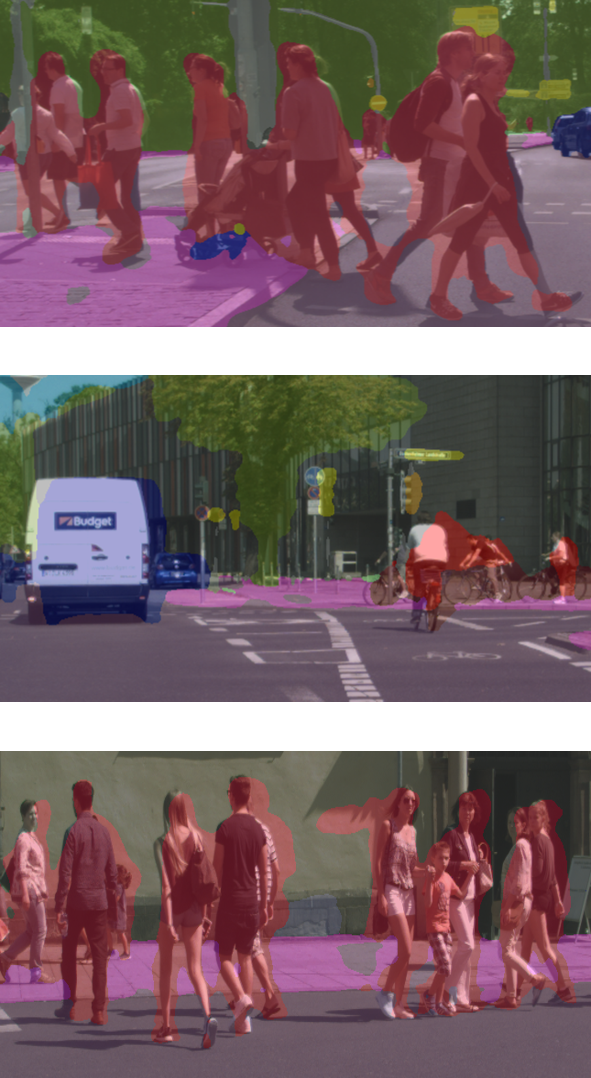}
        \caption{$I_{t+1}$ and $F(I_t)$}
    \end{subfigure}%
    \begin{subfigure}{0.33\textwidth}
        \centering
        \includegraphics[width=0.85\linewidth]{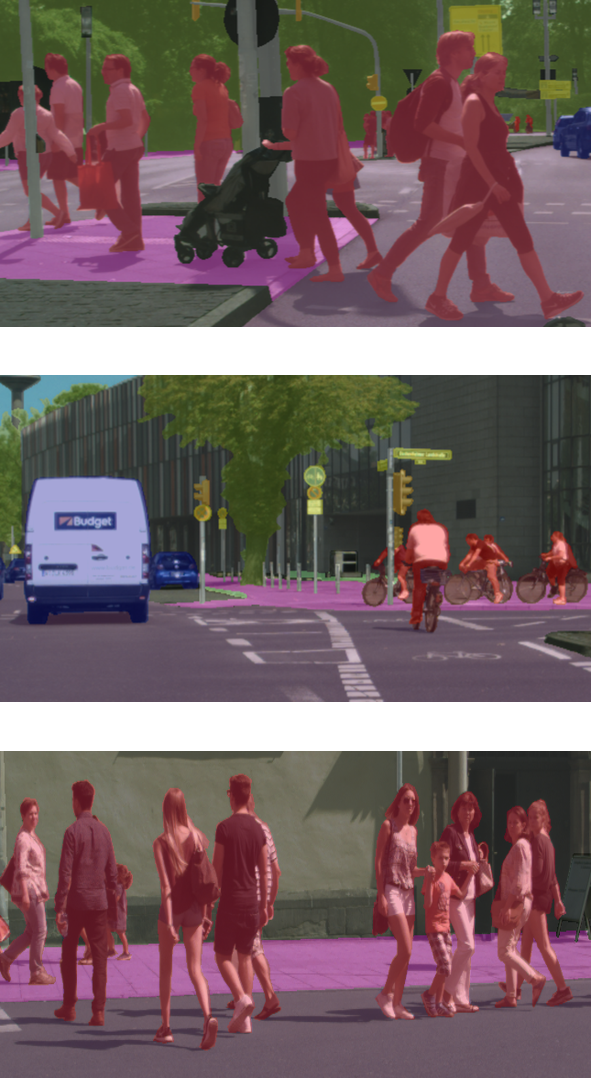}
        \caption{$I_{t+1}$ and $S_{t+1}$ (G.T.)}
    \end{subfigure}%
\caption{Output segmentation of SwiftNet-R18 trained to predict $S_{t+1}$ from $I_t$. We observe anisotropic blur as the network is able to infer some objects movement directions from their orientation.}
\label{flou1}
\vskip -0.2in
\end{figure*}

The results are reported in third line of \cref{bigtable}. We can see a slight but consistent increase of the LAmIoU metric for all networks.
In \cref{flou1}, we show some qualitative results of SwiftNet-R18 overlaying images $I_t$ and $I_{t+1}$.
We notice that the segmentation mask is slightly blurry, as one would expect. However, we note that the blur is often surprisingly anisotropic, \textit{i.e.} the segmentation blur is not surrounding the object, but favours a specific direction.

We conjecture that the network is able to predict some objects' movement based on their orientation. For instance, a person facing left in image $I_t$ is likely to have moved left in the next image  $I_{t+1}$. Similarly, a car facing the camera is more likely to be coming toward the camera, and thus is probably going to look bigger in the following frame.

\subsubsection{Third configuration}

\begin{figure*}[ht]
\centering
    \begin{subfigure}{0.33\textwidth}
        \centering
        \includegraphics[width=0.85\linewidth]{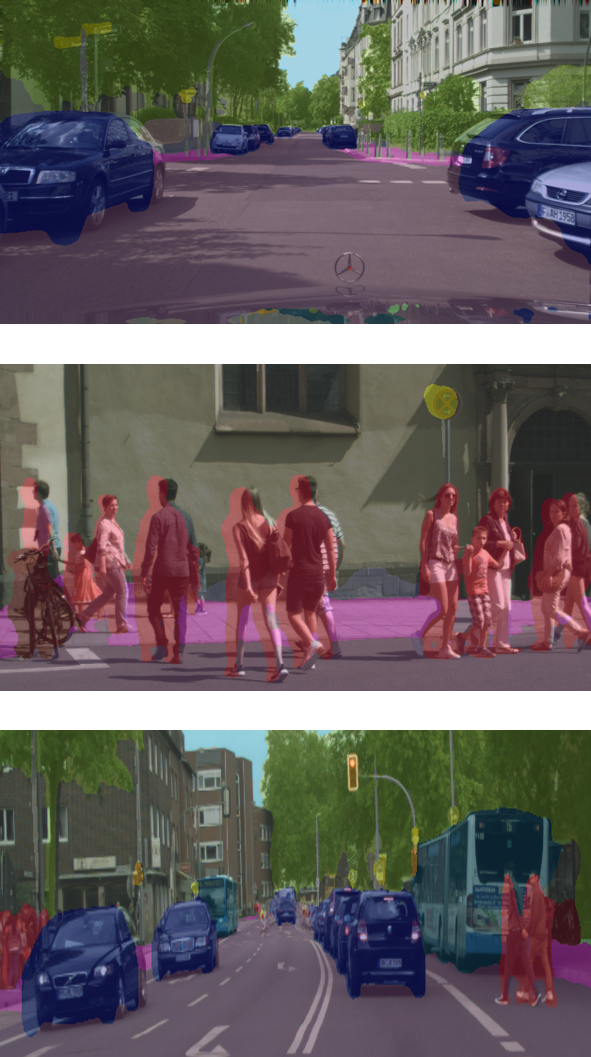}
        \caption{$I_t$ and $F(I_t)$}
    \end{subfigure}%
    \begin{subfigure}{0.33\textwidth}
        \centering
        \includegraphics[width=0.85\linewidth]{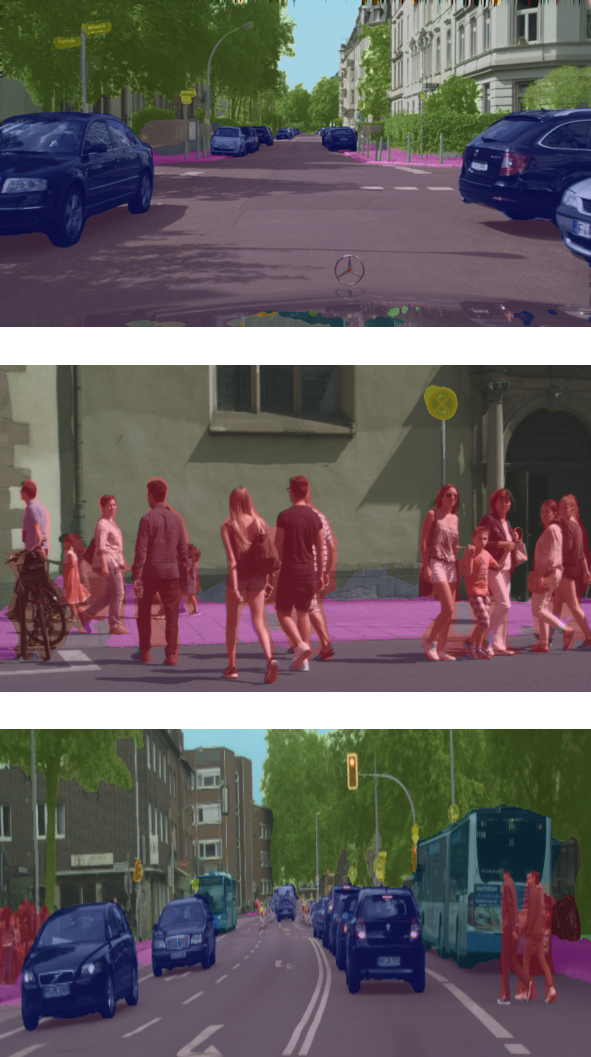}
        \caption{$I_{t+1}$ and $F(I_t)$}
    \end{subfigure}%
    \begin{subfigure}{0.33\textwidth}
        \centering
        \includegraphics[width=0.85\linewidth]{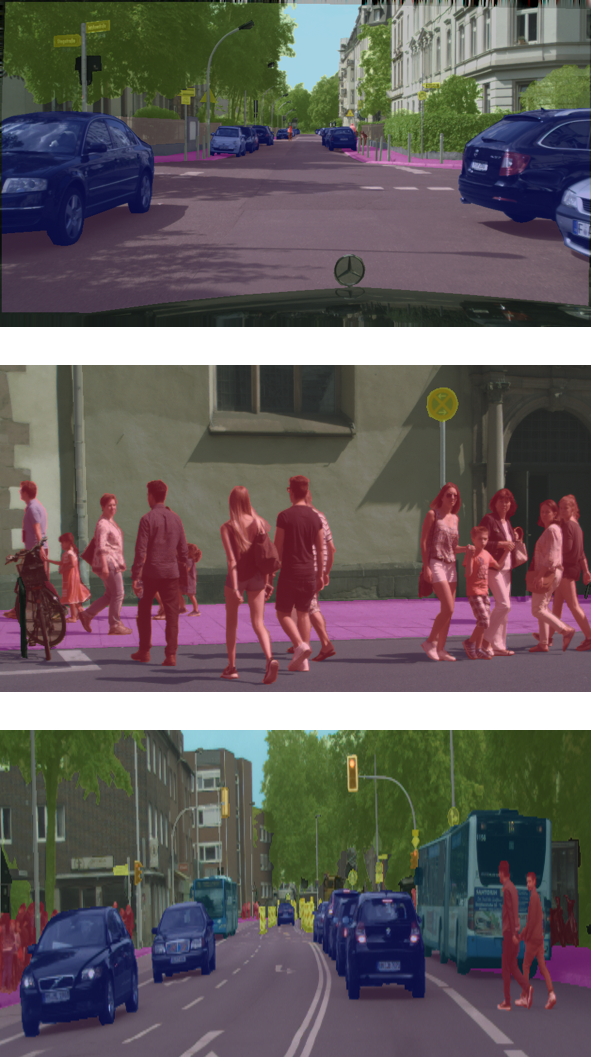}
        \caption{$I_{t+1}$ and $S_{t+1}$ (G.T.)}
    \end{subfigure}%
\caption{Output segmentation of SwiftNet-R18 trained to predict $S_{t+1}$ from $(I_{t-1}, I_t)$. We observe a more precise segmentation as the network has a way to infer relative speeds and directions.}
\label{flou2}
\vskip -0.2in
\end{figure*}

Finally, in order to allow the networks to infer speeds and directions, we train them using both $I_{t-1}$ and $I_t$ as inputs. The training objective remains to predict the future segmentation ground-truth $S_{t+k}$.
Results reported in fourth line of \cref{bigtable} show a consistent improvement of the LAmIoU metric. These numbers confirms the relevance of using  previous frames for our latency-aware objective.
We reported in \cref{flou2} examples of the output segmentation of SwiftNet-R18 overlaid on image $I_t$ and $I_{t+1}$ where it is clear that using an additional input is useful to produce sharper and more accurate segmentation maps.

For practical applications, we have seen that it is relevant to consider a future ground-truth as objective. When doing so, we need to change our training objective, and the result of this last configuration shows that it becomes necessary to use previous frames to get better predictions.
While this result may not be surprising, the great majority of real-time scene segmentation works only use the current input when designing and training their networks. This last result emphasises the fact that networks constructed for real-time contexts should use previous frames to better predict a future target.

\begin{table*}[ht]
\caption{Results for the 3 training configurations for the 4 networks. The first line gives the instant mIoU, the next three lines reports value of the LAmIoU metric for different training configurations. The last three lines give information about latency, frame offset used for the network as explained in \cref{dataset}, and fps of these networks when using a GTX 1080 Ti. Note that the $k$ temporal offset depends on the network, hardware and on the dataset framerate: this offset is greater and leads to poorer performance for slow processing. }
\vskip 0.15in
\label{bigtable}
\centering
\begin{adjustbox}{width=.97\textwidth}
\begin{tabular}{r@{}lll|llll}
\cmidrule[\heavyrulewidth]{2-8}
\multirow{8}{*}{} & Input   & Target (train) & Target (test) & DeepLab-R101  & DeepLab-MN & SwiftNet-R18  & SwiftNet-R18-X  \\
\cmidrule{2-8}
&$I_t$           & $S_t$         & $S_{t}$        & \textbf{0.77} & 0.72       & 0.75          & 0.74        \\               
\multirow{3}{*}{\rotatebox[origin=c]{90}{LAmIoU}} \hspace{0mm} \rdelim\{{3}{20pt} \hspace{-3mm}
&$I_t$                                   & $S_t$         & $S_{t+k}$      & 0.50         & 0.56       & \textbf{0.64} & 0.63          \\
&$I_t$                                   & $S_{t+k}$     & $S_{t+k}$      & 0.53          & 0.57      & \textbf{0.65} & 0.64        \\ 
&$I_{t-1}, I_t$                          & $S_{t+k}$     & $S_{t+k}$      & 0.60          & 0.58       & 0.67          & \textbf{0.69}        \\
\cmidrule{2-8}
\morecmidrules
\cmidrule{2-8}
&\multicolumn{3}{c|}{Frame offset ($k$)}                                  & 4             & 2          & 1             & 1    \\
&\multicolumn{3}{c|}{Latency (ms)}                                        & 195           & 76         & 26            &38   \\
&\multicolumn{3}{c|}{FPS}                                                 & 5             & 13         & 38            & 26   \\
\cmidrule[\heavyrulewidth]{2-8}
\end{tabular}
\end{adjustbox}
\end{table*}

\subsection{Input translations experiment for increased Receptive Field}

When processing simultaneously images from different time-steps $I_{t-1}$ and $I_t$, it is important for the network to have a large receptive field (to be able to map objects from one frame to the other).
To do so without the need of big kernels, we try to offset this load on the input. The idea is to trade part of the computational cost usually associated with the use of big convolutional kernels for the memory cost of having more inputs.

Practically, we concatenate to the current input of the network various translations of the previous image $I_{t-1}$. When introducing translations, we want to compensate for the use of big kernels by allowing the model to simultaneously attend different parts of the image that a normal convolution kernel would not process simultaneously.
Particularly, we change the inputs
$\left\{I_{t-1}, I_t\right\}$
of the previous experiment to $\left\{T_1(I_{t-1}), \cdots,  T_N(I_{t-1}), I_{t-1}, I_t\right\}$.
corresponding to $N$ different fixed translations $T_i$.
The translations offsets were chosen to span a regular grid around the origin.

While initial experiments seemed promising, we further discovered that the only reason for improved LAmIoU was the additional convolutional layer with higher number of kernels that we added at the head of the model to handle the translations where we had replaced the first convolution layer conv(3, 64, $7 \times 7$, $s=2$) 
with the following block:

\[
\begin{array}{l}
  \text{conv}(6 + 3 \times N, 8 \times N, 7\times7, s=2)\\
  \text{BN}(8 \times N)\\
  \text{ReLU}\\
  \text{conv}(8 \times N, 64, 3 \times 3, s=1).
\end{array}
\]
with $N$ the number of translations. Noticing that using additional inputs requires increasing capacity in the early convolutional layers eventually lead to the design of SwiftNet-R18-X presented in \ref{sw_pres} in which we increased the number of kernels in the first two layers.

\section{Conclusion}

We proposed a change in the usual objective of the segmentation task that  makes real-time networks account for their latency when making their predictions.
In addition to providing a new latency-aware ranking, the associated LAmIoU metric is of particular practical relevance as it represents the actual mIoU value one may obtain on a given hardware when taking into account network latency.

We argued the reasons why real-time networks should be latency-aware and we believe introducing a new latency-aware segmentation objective encourages research in anticipatory networks. With this objective in mind, we also proposed addition to networks and training in order to perform better under this new metric.
While the focus of this paper is specifically toward video scene segmentation, the same change of objective is relevant and applicable to a wide range of other ``real-time tasks'' not limited to computer vision.

\section*{Acknowledgements}

We thank Prabhu Teja for his help and relevant remarks on this manuscript. Evann Courdier was supported by the ``Swiss Center for Drones and Robotics - SCDR'' of the Swiss Department of Defence, Civil Protection and Sport via armasuisse S+T under project No 050-38.


\end{document}